\documentclass[runningheads]{llncs}
\usepackage[T1]{fontenc}
\usepackage{graphicx}
\usepackage{booktabs}
\usepackage[misc]{ifsym}
\usepackage{ulem}
\newcommand{\corr}{(\Letter)}
\usepackage{mwe}
\usepackage{multirow}
\usepackage{mhchem}
\usepackage{graphicx} 
\usepackage{float}  
\usepackage{subfigure}
\usepackage{amsmath} 
\usepackage{bm}
\usepackage{amsfonts} 
\usepackage{stmaryrd} 

\usepackage{mathrsfs} 
\usepackage{hyperref}
\hypersetup{hidelinks,
	colorlinks=true,
	allcolors=black,
	pdfstartview=Fit,
	breaklinks=true}
 \usepackage{diagbox}

\usepackage{todonotes}
\usepackage{comment}
 
\usepackage{color}
\definecolor{Orange}{rgb}{1,0.5,0}
\definecolor{Red}{rgb}{1,0,0}
\definecolor{Blue}{rgb}{0,0,1}

\begin{document}

\title{HiGraphDTI: Hierarchical Graph Representation Learning for Drug-Target Interaction Prediction}

\titlerunning{HiGraphDTI for DTI Prediction}


\author{Bin Liu \inst{1} \and
Siqi Wu \inst{1} \and
Jin Wang \inst{1} \thanks{Corresponding author.} \and
Xin Deng \inst{1} \and
Ao Zhou\inst{1}
}

\authorrunning{Bin Liu et al.}

\institute{Key Laboratory of Data Engineering and Visual Computing,\\
Chongqing University of Posts and Telecommunications, China\\
\email{\{liubin, wangjin, dengxin\}@cqupt.edu.cn, \{yunning996, zacqupt\}@gmail.com}
}

\maketitle              

\begin{abstract}
The discovery of drug-target interactions (DTIs) plays a crucial role in pharmaceutical development. 
The deep learning model achieves more accurate results in DTI prediction due to its ability to extract robust and expressive features from drug and target chemical structures.
However, existing deep learning methods typically generate drug features via aggregating molecular atom representations, ignoring the chemical properties carried by motifs, i.e., substructures of the molecular graph. 
The atom-drug double-level molecular representation learning can not fully exploit structure information and fails to interpret the DTI mechanism from the motif perspective.
In addition, sequential model-based target feature extraction either fuses limited contextual information or requires expensive computational resources. 
To tackle the above issues, we propose a hierarchical graph representation learning-based DTI prediction method (HiGraphDTI). 
Specifically, HiGraphDTI learns hierarchical drug representations from triple-level molecular graphs to thoroughly exploit chemical information embedded in atoms, motifs, and molecules.
Then, an attentional feature fusion module incorporates information from different receptive fields to extract expressive target features.
Last, the hierarchical attention mechanism identifies crucial molecular segments, which offers complementary views for interpreting interaction mechanisms.
The experiment results not only demonstrate the superiority of HiGraphDTI to the state-of-the-art methods, but also confirm the practical ability of our model in interaction interpretation and new DTI discovery.  



\end{abstract}

\section{Introduction}

Nowadays, pharmaceutical scientists still rely on existing drug-target interactions (DTIs) to develop novel drugs~\cite{Bagherian_Sabeti_Wang_Sartor_Nikolovska-Coleska_Najarian_2021}. Therefore, there is a pressing need to accurately and efficiently discover new DTIs. Although traditional \textit{in vitro} wet-lab verification can obtain reliable DTIs, the complex experimental process consumes considerable time and labor, making it challenging to screen through a large number of candidates rapidly~\cite{zitnik2019machine}. 
The computational methods receive considerable focus, since they can significantly diminish the resources for screening by predicting reliable DTI candidates~\cite{Abbasi_Razzaghi_Poso_Ghanbari-Ara_Masoudi-Nejad_2021}.
Deep learning models have achieved superior performances in DTI prediction, due to their ability to extract robust and high-quality features from abundant drug and target structure information \cite{sun2020graph,bagherian2021machine}. 
Deep learning DTI prediction methods typically extract drug and target features from their chemical structures and integrate them to infer unseen interactions \cite{Zhao_Yang_Cheng_Li_Wang_2022}. 


Drugs are chemical molecules, represented by either the Simplified Molecular Input Line Entry System (SMILES) strings~\cite{Anderson_Veith_Weininger_1987} or molecular graphs~\cite{10.1093/bioinformatics/btaa921}. 
Convolutional Neural network (CNN)~\cite{10.1093/bioinformatics/btab715} and Transformer~\cite{Huang_Xiao_Glass_Sun_2021,9965612} are utilized to generate drug embeddings via encoding sequential molecular information in SMILES strings.
On the other hand, the molecular graphs explicitly depict atom relations in 2-dimensional geometric space, enabling graph neural networks (GNNs) to extract more informative drug representations~\cite{Hua_Song_Feng_Wu_Kittler_Yu_2022,10.1093/bioinformatics/btac377}. 
Motifs, molecular subgraphs composed of part of atoms and their bonds, usually carry indicative information about the important molecular properties and functions \cite{zhang2021motif}.
Nevertheless, existing GNN-based deep learning models typically learn atom node embeddings and aggregate them via readout or attention-weighted summation to derive molecular representations, ignoring important functional characteristics expressed by motifs. 
Furthermore, current DTI prediction methods only offer the contribution of each atom to the interaction between drug and target, failing to investigate the biological interpretation of DTIs from the motif perspective.

For targets, DTI prediction methods use sequential models, such as CNN~\cite{Tsubaki_Tomii_Sese_2019,Hua_Song_Feng_Wu_Kittler_Yu_2022}, RNN~\cite{Gao_Fokoue_Luo_Iyengar_Dey_Zhang_2018} and Transformer~\cite{9425008} to extract high-level features from their amino acid sequences. 
They commonly select the last layer of the deep neural network as final representations. 
However,  CNN and RNN-based target features lack a broad receptive field~\cite{Tsubaki_Tomii_Sese_2019,Hua_Song_Feng_Wu_Kittler_Yu_2022,Gao_Fokoue_Luo_Iyengar_Dey_Zhang_2018}.
Although Transformer-based target representations fuse every amino acid embeddings, they suffer expensive computational costs~\cite{9425008}. 



In this study, we propose a hierarchical graph representation learning-based DTI prediction method (HiGraphDTI) to enrich the information involved in drug and target features and enhance the interpretation of DTI mechanisms. 
First, we employ hierarchical molecular graph representation to extract atom, motif, and global-level embeddings, enabling atomic information aggregation more orderly and reliable while incorporating more chemical properties. 
Then, we develop an attentional target feature fusion module, which extends receptive fields to improve the expressive ability of protein representations. 
Finally, we design a hierarchical attention mechanism to capture the various level correlations between drugs and targets, providing comprehensive interpretations of DTIs from multiple perspectives. 
Experimental results on four benchmark DTI datasets illustrate that HiGraphDTI surpasses six state-of-the-art methods. The effectiveness of our method in providing valuable biological insights is verified via case studies on multi-level attention weight visualizations.

\section{Related Work}
Predicting drug-target interactions (DTIs) is a crucial area of research in drug development. In recent years, predominant computational approaches comprise two categories: traditional machine learning and deep learning.


Traditional machine learning DTI prediction methods typically rely on manually crafted features, e.g., molecular descriptors for drugs and structural and physicochemical property-based protein features~\cite{sachdev2019comprehensive}.
In~\cite{Jacob2008ProteinligandIP}, the SVM classifier utilizes different kernel functions to determine the similarity of compounds and proteins, and combines chemical and genomic spaces via tensor products. 
EBiCTR~\cite{pliakos2020drug} is an ensemble of bi-clustering trees trained on the reconstructed output space and dyadic
(drug and target) feature space.


Deep learning approaches can alleviate the issue by their capability to learn feature representations. 
DeepDTA~\cite{10.1093/bioinformatics/bty593} only leverages the sequence information of drugs and targets to predict drug-target binding affinity.
DeepConv-DTI~\cite{10.1371/journal.pcbi.1007129} employs convolution on amino acid subsequences of varying lengths to capture local residue patterns in proteins, enriching the information of target features.
TransformerCPI~\cite{Chen_Tan_Wang_Zhong_Liu_Yang_Luo_Chen_Jiang_Zheng_2020} let target features serve as the output of the Transformer encoder and the drug features serve as the input to the Transformer decoder to catch the interactions between drugs and targets. 
MolTrans~\cite{Huang_Xiao_Glass_Sun_2021} introduces a Frequent Consecutive Sub-sequence (FCS) mining algorithm, which utilizes unlabeled data to learn contextual semantic information in SMILES strings and amino acid sequences. The FCS algorithm enhances the expressive power of the model and makes progress in exploiting other information. However, it merely identifies patterns in SMILES strings, which may not correspond to the structural characteristics of drugs. 
IIFDTI~\cite{10.1093/bioinformatics/btac485} comprises four feature components: target features extracted by convolutional neural networks, drug features extracted by graph attention networks, and two interaction features obtained from the Transformer. Similar to MoLTrans, it also incorporates semantic information of SMILES and amino acid sequences. 
DrugBAN~\cite{Bai2023} utilizes a bilinear attention network module to capture local interactions between drugs and targets for DTI prediction. 

Although the aforementioned methods have achieved excellent performance, they still encounter issues: 1. They explore the structural information in drug molecules inadequately. 2. They typically employ summation or averaging as the READOUT function, leading to an unordered aggregation of information in this process. 3. They lack multi-level biological interpretation.

\section{Method}
\begin{figure}[b]
    \centering
    \includegraphics[width=\textwidth]{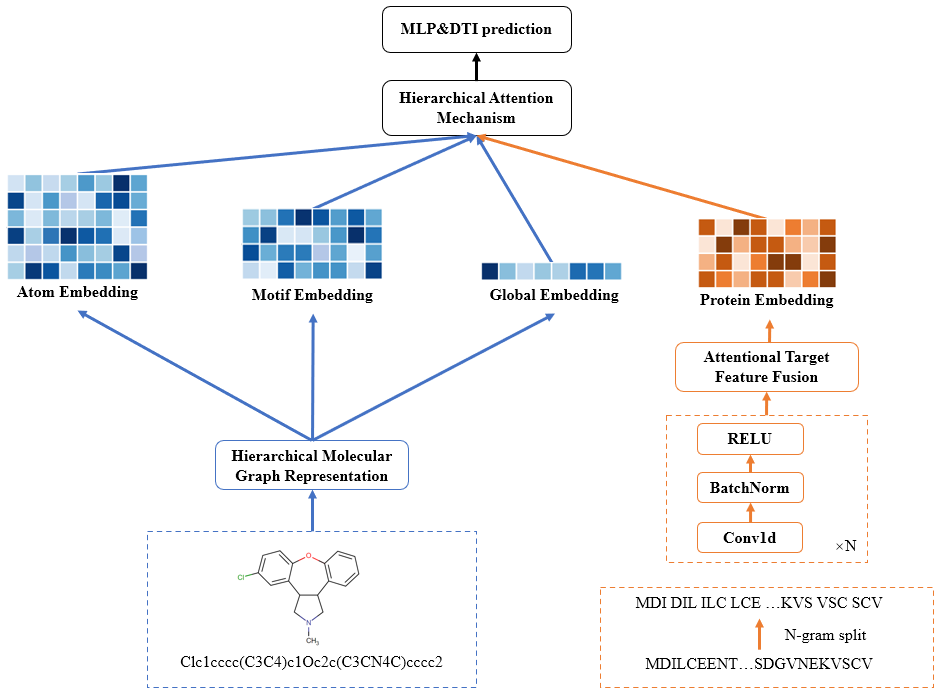} 
    \caption{The overview architecture of HiGraphDTI.
    }
    \label{fig: overview of HiGraphDTI}
\end{figure}

In this section, we illustrate the proposed HiGraphDTI model that predicts interactions between drugs and targets via hierarchical graph representation learning. Fig. \ref{fig: overview of HiGraphDTI} outlines the architecture of HiGraphDTI, which consists of three main modules:
\begin{itemize}
    \item Hierarchical molecular graph representation that extracts drug features, enriching the chemical structure properties and characteristics exploitation and making the information aggregation process more orderly and reliable.
    \item Attentional target feature fusion that adopts a broader receptive field to protein sequence representation extraction.
    \item Hierarchical attention mechanism that captures the correlations between drug and target features from various perspectives, providing comprehensive explanations for DTI mechanisms.
\end{itemize}    


\subsection{Hierarchical Molecular Graph Representation}
\begin{figure}
\centering
    \includegraphics[width=\textwidth]{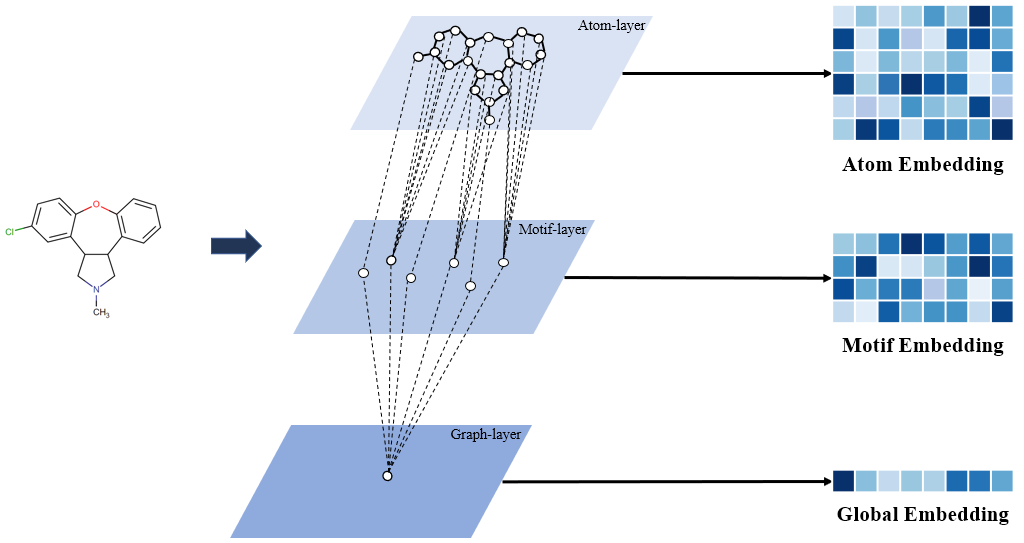} 
    \caption{Hierarchical graph representation construction. In the diagram, solid lines represent bidirectional edges, and dashed lines represent unidirectional edges. 
    }
    \label{fig: Hierarchical graph representation construction}
\end{figure}

Hierarchical graph representation for drugs contains two parts: hierarchical graph construction and message-passing. 
The molecular graph partition process is illustrated in Fig. \ref{fig: Hierarchical graph representation construction}. 

First, We transform the original drug molecules into a graph $G = (V, E)$, where each atom corresponds to a node $v \in V$, and the bonds between atoms correspond to bidirectional edges in $E$. $G$ is the atom layer of the molecular graph.
Then, we divide the drug molecules into multiple functional fragments using the Breaking of Retrosynthetically Interesting Chemical Substructures (BRICS) algorithm, which defines 16 rules and breaks strategic bonds in a molecule that match a set of chemical reactions~\cite{Degen2008}. Following the work~\cite{zhang2021motif}, we supplement an additional partition rule, i.e., disconnecting cycles and branches around minimum rings, to BRICS algorithm to get rid of excessively large fragments.

These obtained fragments, referred to as motifs, construct the second level of the molecular graph. We create a node for each motif, and the collection of nodes is defined as $V_m$. We connect each motif node with its involved atoms in the atom layer, and the collection of these edges is defined as $E_m$. 
To avoid the over-smoothing issue in graph neural networks and make message aggregation more reasonable, these edges are unidirectional, pointing from the atom layer to the motif layer. 

Finally, to aggregate the global information of drug molecules, we construct a global node $V_g$, which is the graph-layer. We establish connections between it and all motif nodes, and the collection of these edges is referred to as $E_g$. These edges are also unidirectional, pointing from motif nodes to the global node $V_g$. 
The final hierarchical graph is constructed as follows:
\begin{equation}
\bar{G} = (\bar{V}, \bar{E}), \bar{V} = (V, V_m, V_g), \bar{E} = (E, E_m, E_g)
\end{equation}

Given the triple-layer molecular graph, we employ Graph Isomorphism Network (GIN) to propagate messages and learn node embeddings due to its superior expressive power demonstrated by  Weisfeiler-Lehman (WL) test~\cite{DBLP:journals/corr/abs-1810-00826}.
Specifically, the message-passing formula of GIN is:
\begin{equation}
\mathbf{h}_{v}^{l} = MLP^{l}(\mathbf{h}_{v}^{l-1} + \sum_{u\in \mathcal{N}(v)}(\mathbf{h}_{v}^{l-1} + \mathbf{W}^l\mathbf{X}_{uv}))
\end{equation}
where $MLP^{l}$ represents a multi-layer perceptron (MLP) for node features utilized to update nodes, $\mathbf{X}_{uv}$ represents the edge embeddings between nodes $u$ and $v$, $\mathbf{W}^l$ represents the embedding parameters of $\mathbf{X}_{uv}$ for each layer and $\mathbf{h}_{v}^{0} = \mathbf{X}_{v}$ is the input node feature of $v \in \bar{V}$. After multiple iterations of updates, we obtain the final embeddings of atom, motif, and global nodes, denoted as $\mathbf{H}_{a} \in \mathcal{R}^{|a|\times d}$, $\mathbf{H}_{m} \in \mathcal{R}^{|m|\times d} $ and $\mathbf{H}_{g} \in \mathcal{R}^{1\times d}$ respectively, where $|a|$ is the number of atoms, $|m|$ is the number of motifs. We adopt $\mathbf{H}_{g}$ as the representation of the whole drug molecule.

\subsection{Attentional Target Feature Fusion}


Following previous work~\cite{Tsubaki_Tomii_Sese_2019}, we partition the target sequence into 3-gram amino acids to obtain the initial vector, denoted as $\mathbf{X}_{P} = \{x_1, x_2, ..., x_l\}$, where $x_i \in \mathcal{R}^d$ represents the embedding of the $i$-th segment, $l$ is the number of the partitioned sequences, and $d$ is the embedding dimension. To better aggregate critical features in the protein vector representation, We design a one-dimensional (1D) convolutional neural network with layer-wise decreasing channels, and the formula for each layer is as follows:
\begin{equation}
\mathbf{X}_i = Relu(BN_i(Conv1D_i(\mathbf{X}_{i-1})))
\end{equation}
where $\mathbf{X}_i$ represents the feature representation for the $i$-th layer, and $\mathbf{X}_0 = \mathbf{X}_P$, $Conv1D$ represents the 1D convolution in the $i$-th layer with the kernel size of 15 and the output channels reduced by half, $Relu$ represents the ReLu nonlinear activation function, $BN_i$ represents the batch normalization in the $i$-th layer.

\begin{figure}
    \centering
    \includegraphics[width=0.8\textwidth]{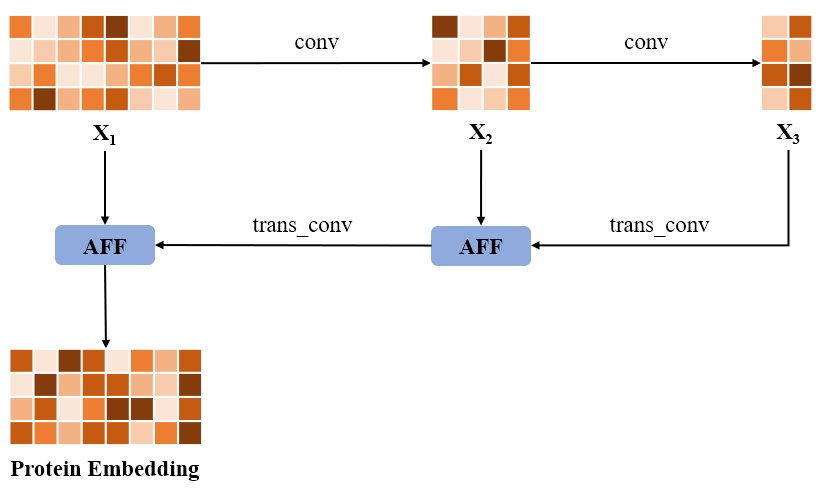} 
    \caption{The overview architecture of feature fusion module for protein. The high-level feature is mapped to the same dimension as the low-level using transposed convolution and then input into the AFF module for fusion. Taking the result as high-level features, repeat the operation.
    }
    \label{fig: The overview architecture of feature fusion module for protein}
\end{figure}

We obtain target feature representations $\mathbf{X}_1$, $\mathbf{X}_2$, $\mathbf{X}_3$ at three different convolutional layers. To aggregate target information, we adapt the attentional feature fusion (AFF) module~\cite{Dai_Gieseke_Oehmcke_Wu_Barnard_2021} tailored to amino acid sequences. 
The process is depicted in Fig. \ref{fig: The overview architecture of feature fusion module for protein}. We perform transposed convolution on $\mathbf{X}_3$ to map it to the same dimension as $\mathbf{X}_2$ and then put it into the AFF module. Next, Map the result to the same dimension as $\mathbf{X}_1$ and put it into the AFF module to obtain the outcome, denoted as $\mathbf{H}_P \in \mathcal{R}^{l \times d}$, where $l$ is the number of partitioned sequences, is the embedding dimension. 

\begin{figure}
    \centering
    \includegraphics[width=0.7\textwidth]{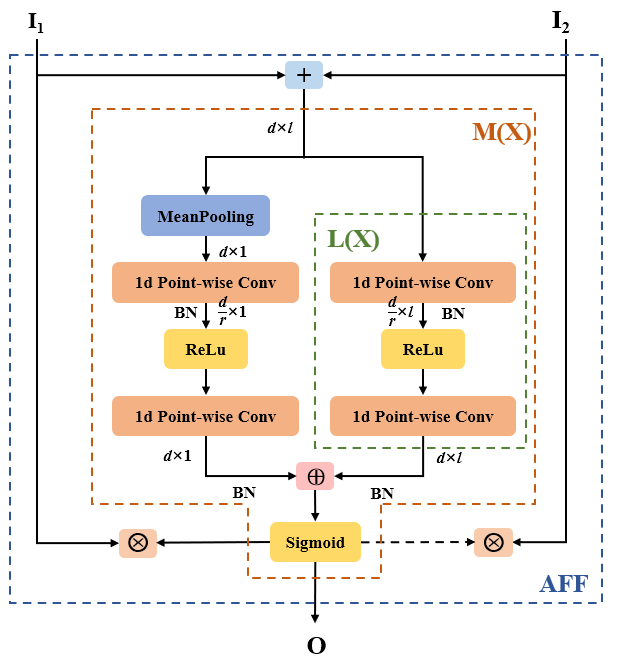} 
    \caption{The architecture of AFF module, which utilizes operation $M$ to compute the attention matrix for weighted aggregation of inputs $\mathbf{I}_1$ and $\mathbf{I}_2$.}
    \label{fig: The architecture of AFF module}
\end{figure}

The detailed illustration of AFF module are shown in Fig. \ref{fig: The architecture of AFF module}. AFF module receives two inputs, $\mathbf{I}_1$ and $\mathbf{I}_2$, where $\mathbf{I}_1$ is the high-level feature after transposed convolution, and $\mathbf{I}_2$ is the low-level feature. We combine the information from those through element-wise summation as $\mathbf{I}$ and feed the result into module $M$ to extract additional information. Module $M$ achieves channel attention across multiple scales by changing the spatial pooling size. It mainly consists of two parts: one for extracting global features and the other for extracting local features, as illustrated in Eq. \eqref{eq:M}.
\begin{equation}
M(\mathbf{I}) =  \sigma(L(\mathbf{I}) \oplus L(MeanPooling(\mathbf{I})))
\label{eq:M}
\end{equation}
where $\sigma$ is the Sigmoid function, $MeanPooling(\mathbf{I}) = \frac{1}{l} \sum_{i=1}^{l} \mathbf{I}[i,:]$ is the average pooling along columns, $\oplus$ refers to the broadcasting addition and $L(\mathbf{I})$ is defined as:
\begin{equation}
L(\mathbf{I}) =  BN(PWConv_2(Relu(BN(PWConv_1(\mathbf{I})))))
\end{equation}
In L($\mathbf{I}$), $PWConv_{1}$ and $PWConv_{2}$ refer to two point-wise 1D convolutions to capture information from diverse channels and maintain the model as lightweight as possible. 
After applying the module $M$, We obtain the attention matrix.
To get the final output, we perform the following operations.
\begin{equation}
\mathbf{O} = M(\mathbf{I}) \otimes \mathbf{I}_1 + (1 - M(\mathbf{I})) \otimes \mathbf{I}_2
\end{equation}
where $\otimes$ denotes the element-wise multiplication. In Fig. \ref{fig: The architecture of AFF module}, the black dashed line denotes $(1-M(\mathbf{I}))$. $M(\mathbf{I})$ and $1-M(\mathbf{I})$ are real arrays in the range of 0 to 1, facilitating a weighted sum of $\mathbf{I}_1$ and $\mathbf{I}_2$.

\subsection{Hierarchical Attention Mechanism}

We design a hierarchical attention mechanism to capture the correlation between triple-level drug features ($\mathbf{H}_a$, $\mathbf{H}_m$, $\mathbf{H_g}$) and target features ($\mathbf{H}_P$). 
Graph level drug features $\mathbf{H}_g$ aggregates the global molecular information after cross-level message-passing. It consists of only a single vector, lacking robustness.
Incorporating additional information could significantly decrease its express capacity. 
Therefore, we only incorporate drug features into the target embedding $\mathbf{H}_P$. We calculate the attention between targets and different levels of drugs using the following formula:

\begin{equation}
\mathbf{Attn}_a = Relu(\mathbf{H}_P^{\top}\mathbf{W}_a\mathbf{H}_a)
\end{equation}

\begin{equation}
\mathbf{Attn}_m = Relu(\mathbf{H}_P^{\top}\mathbf{W}_m\mathbf{H}_m)
\end{equation}

\begin{equation}
\mathbf{Attn}_g = Relu(\mathbf{H}_P^{\top}\mathbf{W}_g\mathbf{H}_g)
\end{equation}
where $\mathbf{Attn}_a \in \mathcal{R}^{l\times|a|}$, $\mathbf{Attn}_m\in \mathcal{R}^{l\times|m|}$, $\mathbf{Attn}_g\in \mathcal{R}^{l\times 1}$ represent attention matrices between protein partitioned sequence and different levels (atom, motif, and global) of the drug molecule. Next, we calculate the mean along the rows for each attention matrix, resulting in three attention vectors $\mathbf{A}_a$, $\mathbf{A}_m$, $\mathbf{A}_g$. The summation of these vectors utilized as weights for updating $\mathbf{H}_P$ yields the protein representation enriched with drug information. Its formula is as follows: 
\begin{equation}
\mathbf{F}_P = \mathbf{H}_P\cdot(SF(\mathbf{A}_a) + SF(\mathbf{A}_m) + SF(\mathbf{A}_g))
\end{equation}
where $SF$ is the softmax function.

Finally, we concatenate $\mathbf{H}_g$ and $\mathbf{F}_P$ and then feed them into a multi-layer perceptron (MLP) model to derive the probability of drug-target interaction $\mathbf{\hat{Y}}$. The binary cross-entropy loss is utilized for training our model.
\begin{equation}
\mathcal{L} = -\frac{1}{N}\sum_{i}^{N} y_i\cdot log(\hat{y_i} ) + (1-y_i)\cdot log(1-\hat{y_i})
\end{equation}
where $y_i$ is the true label, $\hat{y_i}$is the predicted label, $N$ is the number of training samples.

\section{Experiments}

\subsection{Experimental Setup} 
We select four benchmark datasets in the DTI field to evaluate our model, including Human dataset~\cite{Liu_Sun_Guan_Zheng_Zhou_2015}, Caenorhabditis elegant (C.elegans) dataset~\cite{Liu_Sun_Guan_Zheng_Zhou_2015}, BindingDB dataset~\cite{Gao_Fokoue_Luo_Iyengar_Dey_Zhang_2018}, GPCR dataset~\cite{Chen_Tan_Wang_Zhong_Liu_Yang_Luo_Chen_Jiang_Zheng_2020}. Human and C.elegans datasets are created using a systematic screening framework to obtain highly credible negative samples~\cite{Liu_Sun_Guan_Zheng_Zhou_2015}. GPCR dataset is constructed through the GLASS database~\cite{Chan_Zhang_Yang_Brender_Hur_Özgür_Zhang_2015}, which uses scores to describe the drug-target affinity(DTA). To obtain samples for DTIs, GPCR uses a threshold of 6.0 to categorize positive and negative samples. The BindingDB dataset~\cite{Gao_Fokoue_Luo_Iyengar_Dey_Zhang_2018} primarily focuses on the interactions of small molecules, and it is well-divided into non-overlapping training, validation, and test sets. 
Table \ref{table:datasets} presents the statistics of the mentioned datasets.

\begin{table}[hbt]
\centering
\setlength{\tabcolsep}{5pt}
\caption{Statistics of datasets.}
\label{table:datasets}
\begin{tabular}{@{}cccccc@{}}
\toprule[1pt]
Datasets  & Targets & Drugs & Interactions & Positive & Negative \\ 
\midrule
Human     & 852     & 1052   & 6738     & 3369     & 3369     \\
C.elegans & 2504    & 1434   & 8000     & 4000     & 4000     \\
BindingDB & 812     & 49745  & 61258    & 33772    & 27486    \\
GPCR      & 356     & 5359   & 15343    & 7989     & 7354     \\ 
\bottomrule[1pt]
\end{tabular}

\end{table}
For the Human and C.elegans datasets, we employ a five-fold cross-validation approach. They are divided into training set, validation set and test set according to the ratio of 8:1:1. For the BindingDB dataset,  the training set, validation set and test set are partitioned well~\cite{Gao_Fokoue_Luo_Iyengar_Dey_Zhang_2018}.
For the GPCR dataset, the training set and test set are divided well~\cite{Chen_Tan_Wang_Zhong_Liu_Yang_Luo_Chen_Jiang_Zheng_2020}. We randomly select the $20\%$ of the training set as the validation set.

We select six state-of-the-art DTI prediction methods for comparison: DeepDTA \cite{10.1093/bioinformatics/bty593}, DeepConv-DTI \cite{10.1371/journal.pcbi.1007129}, MolTrans \cite{Huang_Xiao_Glass_Sun_2021}, TransformerCPI \cite{Chen_Tan_Wang_Zhong_Liu_Yang_Luo_Chen_Jiang_Zheng_2020}, IIFDTI \cite{10.1093/bioinformatics/btac485}, and DrugBAN \cite{Bai2023}. A brief introduction to the methods mentioned above is provided in the Supplementary Materials. To adapt DeepDTA, a drug-target affinity prediction model, to the DTI prediction task, we replace the loss function in its last layer with binary cross-entropy loss.

We choose four metrics for evaluating our models: AUC (the area under the receiver operating characteristic curve), AUPR (the area under the precision-recall curve), Precision, and Recall. We execute all models ten times using different random seeds, calculating their averages to compare performance.  

For all datasets, we save the model parameters that achieve the highest AUC on the validation set. Then, we evaluate its performance on the test set to obtain results. For each dataset, we execute the experiments ten times with different ten seeds and calculate their average and standard deviation (std) as the final results to compare. Details regarding dataset partitioning and model hyperparameter settings are available in the Supplementary Materials.
The codes of our model are available at \href{https://anonymous.4open.science/r/HiGraphDTI-08FB}{https://anonymous.4open.science/r/HiGraphDTI-08FB}.


\subsection{Comparison Results}
As shown in Table \ref{table:AUC_result} and Table \ref{table:AUPR_result}, HiGraphDTI outperforms the six baselines in terms of AUC and AUPR on all datasets. We attribute the excellent performance to three merits of HiGraphDTI. First, using hierarchical graph representation allows drugs to aggregate information across different levels, enriching the molecular structure representation. Second, employing feature fusion modules enables targets to capture information from different receptive fields, enhancing the protein sequence representation. Thrid, Applying hierarchical attention mechanisms computes interactive attention between different levels of drugs and targets, augmenting the interaction information between drugs and targets.

IIFDTI ranks second on the Human, C. elegans and GPCR datasets. The innovation of IIFDTI lies in its utilization of Word2Vec to separately extract feature representations from SMILES and amino acid sequences. It incorporates textual information encoded in SMILES, while HiGraphDTI enriches hierarchical information in molecular graph representations. Compared to compressed textual information, hierarchically aggregated information based on molecular chemical properties is more expressive. At the same time, after hierarchical partitioning, our method can calculate attention scores between different levels and the target. That enriches the information of interaction features and allows for diverse biological interpretations at different levels of DTI. HiGraphDTI surpasses IIFDTI in AUC and AUPR, especially on the GPCR dataset, with improvements of $1.3\%$ and $0.8\%$, respectively. 
For the larger dataset BindingDB, DrugBAN is the second-best in terms of AUC. DrugBAN utilizes graph neural networks and convolutional neural networks to extract feature representations for drugs and targets. It employs its proposed Bilinear Attention Network to obtain interaction features. However, It does not incorporate additional information to enrich its feature representation, resulting in its inferiority to HiGraphDTI. Furthermore, HiGraphDTI also exhibits advantages over IIFDTI on the BindingDB dataset, achieving improvements of $0.9\%$ in AUC and $1.1\%$ in AUPR.
The results for precision and recall are presented in the supplementary materials.

\begin{table}
\centering
\setlength{\tabcolsep}{5pt}
\caption{Experiment results in terms of AUC, where the best and runner-up results are highlighted in bold and underlined, respectively.}\label{table:AUC_result}
\begin{tabular}{ccccc} 
\toprule[1pt]
\diagbox{Model}{Dataset} & Human                  & C.elegans              & BindingDB              & GPCR                    \\ 
\toprule
DeepDTA                  & 0.972 (0.001)          & 0.983 (0.001)          & 0.934 (0.007)          & 0.776 (0.006)           \\ 
\toprule
DeepConv-DTI             & 0.967 (0.002)          & 0.983 (0.002)          & 0.922 (0.003)          & 0.752 (0.011)           \\ 
\toprule
MolTrans                 & 0.974 (0.002)          & 0.982 (0.003)          & 0.899 (0.006)          & 0.807 (0.004)           \\ 
\toprule
TransformerCPI           & 0.970 (0.006)          & 0.984 (0.002)          & 0.933 (0.011)          & 0.842 (0.007)           \\ 
\toprule
IIFDTI                   & \uline{0.984 (0.003)}  & \uline{0.991 (0.002)}  & 0.944 (0.003)          & \uline{0.845 (0.008)}   \\ 
\toprule
DrugBAN                  & 0.984 (0.001)          & 0.989 (0.001)          & \uline{0.945 (0.007)}  & 0.837 (0.010)           \\ 
\toprule
Ours                     & \textbf{0.985 (0.001)} & \textbf{0.993 (0.001)} & \textbf{0.954 (0.003)} & \textbf{0.858 (0.004)}  \\
\toprule
\end{tabular}
\end{table}

\begin{table}
\centering
\setlength{\tabcolsep}{5pt}
\caption{Experiment results in terms of AUPR, where the best and runner-up results are highlighted in bold and underlined, respectively.}\label{table:AUPR_result}
\begin{tabular}{ccccc} 
\toprule[1pt]
\diagbox{Model}{Dataset} & Human                  & C.elegans              & BindingDB              & GPCR                    \\ 
\toprule
DeepDTA                  & 0.973 (0.002)          & 0.984 (0.007)          & 0.934 (0.008)          & 0.762 (0.015)           \\ 
\toprule
DeepConv-DTI             & 0.964 (0.004)          & 0.985 (0.001)          & 0.921 (0.004)          & 0.685 (0.010)           \\ 
\toprule
MolTrans                 & 0.976 (0.003)          & 0.982 (0.003)          & 0.897 (0.010)          & 0.788 (0.009)           \\ 
\toprule
TransformerCPI           & 0.974 (0.005)          & 0.983 (0.003)          & 0.934 (0.015)          & 0.837 (0.010)           \\ 
\toprule
IIFDTI                   & \uline{0.985 (0.003)}  & \uline{0.992 (0.003)}  & \uline{0.945 (0.004)}          & \uline{0.842 (0.007)}   \\ 
\toprule
DrugBAN                  & 0.981 (0.001)          & 0.990 (0.002)          & 0.944 (0.005)          & 0.823 (0.013)           \\ 
\toprule
Ours                     & \textbf{0.988 (0.001)} & \textbf{0.993 (0.001)} & \textbf{0.955 (0.003)} & \textbf{0.850 (0.003)}  \\
\toprule
\end{tabular}
\end{table}

\subsection{Ablation Experiment}
\begin{figure}[t]
    \centering
    \includegraphics[width=0.7\textwidth]{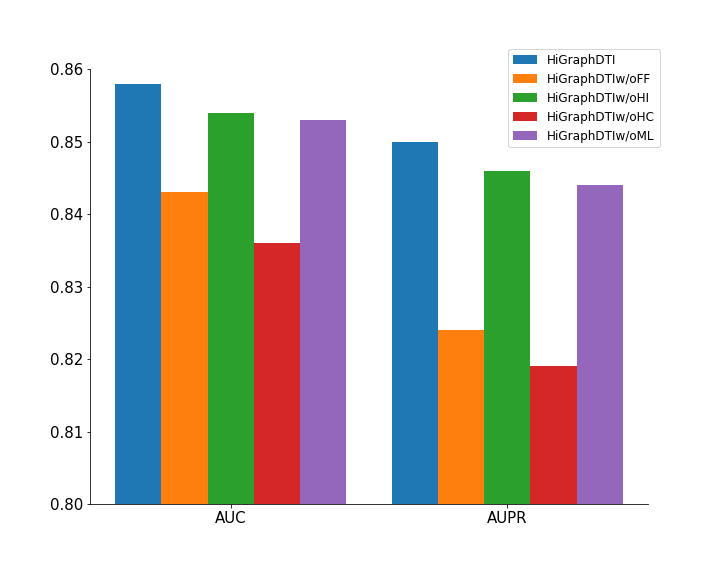} 
    \caption{Ablation experiment results on the GPCR dataset}
    \label{fig: The results of the ablation experiment}
\end{figure}

To validate the effectiveness of each module in HiGraph, we design the following ablation experiments.
\begin{itemize}
    \item HiGraphDTI$w/o$FF: We remove the target feature fusion module and retain the last output convolutional layer as target representation.
    \item HiGraphDTI$w/o$HI: We remove all attentions between drugs and targets. We only concatenate the global-level features of drugs and the mean of target features for prediction.
    \item HiGraphDTI$w/o$HC: We remove the hierarchical structure from the graph representation and only use atom-level embeddings to construct drug features. 
    \item HiGraphDTI$w/o$ML: We remove the motif-level nodes from the hierarchical molecular graph and only utilize atom and global nodes to construct drug features.
\end{itemize}
The experimental results on the GPCR dataset are shown in Fig. \ref{fig: The results of the ablation experiment}. The result of HiGraphDTI$w/o$FF validates the importance of the feature fusion module in constructing target features. Losing multiple receptive fields leads to a decrease in model performance. 
The result of HiGraphDTI$w/o$HI demonstrates the validity of the hierarchical attention mechanisms. It comprehends the interaction between drugs and targets from different perspectives, enhancing the understanding and predictive capability of the model.
Finally, the comparison between HiGraphDTI$w/o$HC and HiGraphDTI$w/o$ML confirms the superiority of hierarchical graph representation learning methods in drug feature extraction. The multi-layered structure enriches the expression of drug features.

\subsection{Attention Interpretation}
\begin{figure}[t]
\centering
\subfigure[Attention Weights for each amino acid of PDB: 1N28]{
\includegraphics[width=\textwidth]{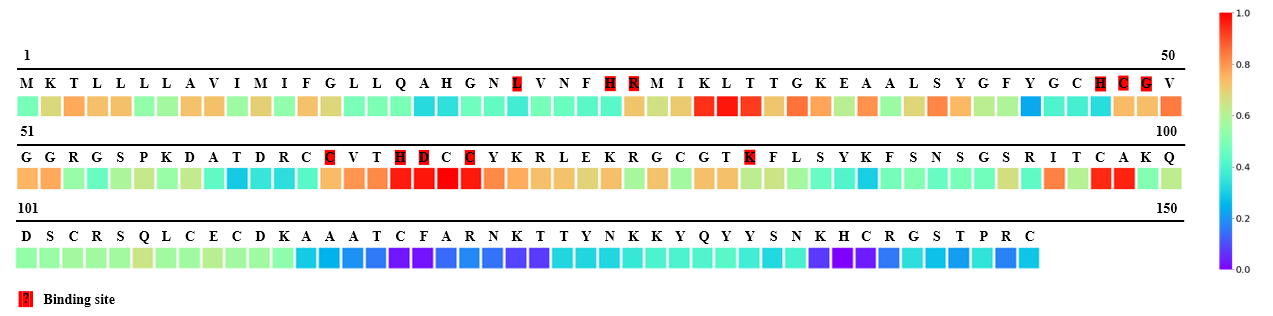}
\label{fig:attention4DTI_target}
}
\subfigure[3D visualization of docking interaction of PDB: I3N with PDB: 1N28]{
\includegraphics[width=0.48\textwidth]{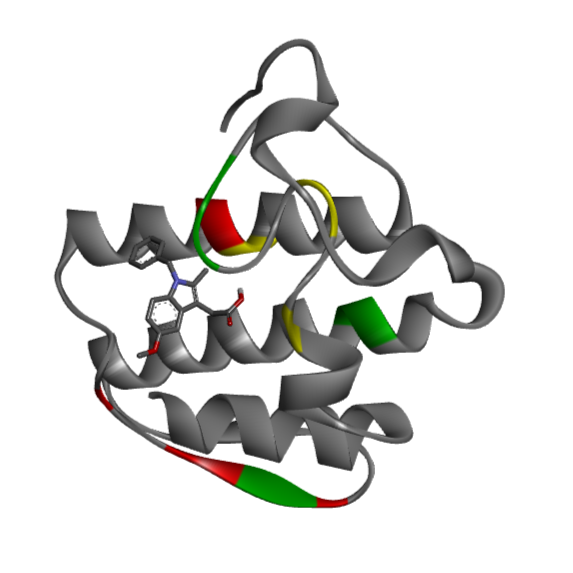}
\label{fig:attention4DTI_3ddocking}
}
\caption{Visualization of target attention weights for interaction of PDB: I3N and PDB: 1N28}
\label{fig:attention4DTI}
\end{figure}

The hierarchical attention mechanism not only enhances model performance but also assists us in understanding the drug-target interaction from various insights. In this part, we utilize the attention weights to interpret the effectiveness of the hierarchical attention mechanism. Furthermore, we illustrate the drug-target interaction from the atom and motif levels to offer valuable assistance for drug discovery.


To better understand the interaction between drug and target, we choose target PDB: 1N28 and drug (ligands) PDB: I3N (\ce{C19H19NO3}) as a case study. We use the hierarchical attention mechanism to calculate the attention vector $\mathbf{B}_P = SF(\mathbf{A}_a) + SF(\mathbf{A}_m) + SF(\mathbf{A}_g) \in \mathcal{R}^l$, which demonstrates the distribution of amino acid attention weights. The values in $\mathbf{B}_P$ are all within the range of 0 to 1. The attention weights for each amino acid of PDB:1N28 are shown in Fig. \ref{fig:attention4DTI_target}, where different colors represent varying attention weights. The actual binding sites are represented by amino acid letters with a red background. 
From Fig. \ref{fig:attention4DTI_target}, we can observe that the model gives high attention to six among the total eleven binding sites. In addition, the model provides seven other positions (located at 30, 31, 32, 69, 70, 97, 98) with high attention weights, which could serve as potential binding sites for future chemical experiments. Fig. \ref{fig:attention4DTI_3ddocking} depicts the 3D visualization of the docking interaction of PDB: I3N and PDB: 1N28, where red regions represent binding sites with high attention weights, yellow segments indicate the binding site with low attention weights, green regions represent the high attention weighted amino acids that have not been recognized as binding sites.

\begin{figure}[t]
    \centering
    \includegraphics[width=0.8\textwidth]{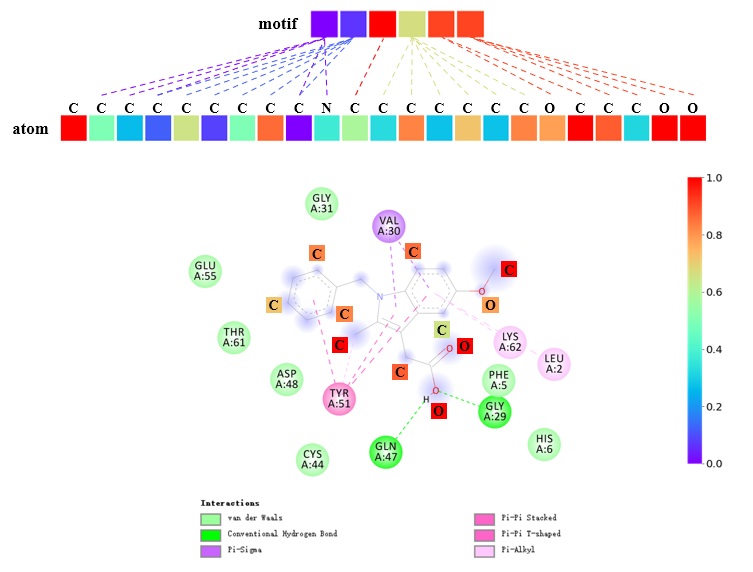} 
    \caption{Visualization of drug attention weights for the interaction of PDB: I3N and PDB:1N28}
    \label{fig: atom_motif}
\end{figure}

In the process of computing $\mathbf{B}_P$, we obtain three attention matrices: $\mathbf{Attn}_a \in \mathcal{R}^{l\times|a|}$, $\mathbf{Attn}_m\in \mathcal{R}^{l\times|m|}$, and $\mathbf{Attn}_g \in \mathcal{R}^{l\times1}$. We further average every column of $\mathbf{Attn}_a$, $\mathbf{Attn}_m$ to obtain the attention vector $\mathbf{B}_a \in \mathcal{R}^{|a|}$, $\mathbf{B}_m \in \mathcal{R}^{|m|}$, where each element illustrates the importance of each node to the interaction. Visualization of drug attention weights for the interaction of PDB: I3N and PDB: 1N28 are shown in Fig. \ref{fig: atom_motif}, where dashed lines of the same color connect motif and its composed atoms. 
There are fifteen atoms interacting with at least one amino acid, where $2/3$ attention weights exceed 0.6. The corresponding motif nodes also exhibit high attention weights. 
It can be observed that the $11$-th atom (\ce{C}) is an active node in the docking simulation. While its atom attention weight is not high, the $3$-rd motif node containing it has a high attention weight, serving as a powerful supplement. This validates that the hierarchical graph representation approach to constructing drug features permits the model to better discern the importance of nodes and ensures crucial nodes are not overlooked during the drug development process.

\section{Conclusion}
In this paper, we propose a novel model to predict DTI, named HiGraphDTI. We utilize hierarchical molecular graph representation to construct drug features, which possess more information about drug structures and a more reasonable way to convey messages. To expand the receptive field of target features, we design the attentional target feature fusion strategy to obtain more informative protein representations. Furthermore, with the hierarchical attention mechanism, we catch the interactive information between drugs and targets from multiple views. To validate the effectiveness of our model, we compare it with six state-of-the-art models on four datasets. The experimental results indicate that our model outperforms comparing baselines in terms of AUC and AUPR metrics. Finally, 
the visualizations of attention weights confirm the interpretation ability of HiGraph to support new drug discovery.

%
%
%
\bibliographystyle{splncs04}
\bibliography{ref}

\begin{thebibliography}{10}
\providecommand{\url}[1]{\texttt{#1}}
\providecommand{\urlprefix}{URL }
\providecommand{\doi}[1]{https://doi.org/#1}

\bibitem{Abbasi_Razzaghi_Poso_Ghanbari-Ara_Masoudi-Nejad_2021}
Abbasi, K., Razzaghi, P., Poso, A., Ghanbari-Ara, S., Masoudi-Nejad, A.: Deep learning in drug target interaction prediction: Current and future perspectives. Current Medicinal Chemistry p. 2100–2113 (Apr 2021)

\bibitem{Anderson_Veith_Weininger_1987}
Anderson, E., Veith, G., Weininger, D.: Smiles: a line notation and computerized interpreter for chemical structures.  (Jan 1987)

\bibitem{Bagherian_Sabeti_Wang_Sartor_Nikolovska-Coleska_Najarian_2021}
Bagherian, M., Sabeti, E., Wang, K., Sartor, M.A., Nikolovska-Coleska, Z., Najarian, K.: Machine learning approaches and databases for prediction of drug-target interaction: a survey paper. Briefings in Bioinformatics p. 247–269 (Jan 2021)

\bibitem{bagherian2021machine}
Bagherian, M., Sabeti, E., Wang, K., Sartor, M.A., Nikolovska-Coleska, Z., Najarian, K.: Machine learning approaches and databases for prediction of drug-target interaction: a survey paper. Briefings in bioinformatics  \textbf{22}(1),  247--269 (2021)

\bibitem{Bai2023}
Bai, P., Miljkovi{\'{c}}, F., John, B., Lu, H.: Interpretable bilinear attention network with domain adaptation improves drug--target prediction. Nature Machine Intelligence  \textbf{5}(2),  126--136 (Feb 2023)

\bibitem{Chan_Zhang_Yang_Brender_Hur_Özgür_Zhang_2015}
Chan, W.K.B., Zhang, H., Yang, J., Brender, J.R., Hur, J., Özgür, A., Zhang, Y.: Glass: a comprehensive database for experimentally validated gpcr-ligand associations. Bioinformatics p. 3035–3042 (Sep 2015)

\bibitem{Chen_Tan_Wang_Zhong_Liu_Yang_Luo_Chen_Jiang_Zheng_2020}
Chen, L., Tan, X., Wang, D., Zhong, F., Liu, X., Yang, T., Luo, X., Chen, K., Jiang, H., Zheng, M.: Transformercpi: improving compound–protein interaction prediction by sequence-based deep learning with self-attention mechanism and label reversal experiments. Bioinformatics p. 4406–4414 (Aug 2020)

\bibitem{9425008}
Cheng, Z., Yan, C., Wu, F.X., Wang, J.: Drug-target interaction prediction using multi-head self-attention and graph attention network. IEEE/ACM Transactions on Computational Biology and Bioinformatics  \textbf{19}(4),  2208--2218 (2022)

\bibitem{10.1093/bioinformatics/btac485}
Cheng, Z., Zhao, Q., Li, Y., Wang, J.: {IIFDTI: predicting drug–target interactions through interactive and independent features based on attention mechanism}. Bioinformatics  \textbf{38}(17),  4153--4161 (07 2022)

\bibitem{Dai_Gieseke_Oehmcke_Wu_Barnard_2021}
Dai, Y., Gieseke, F., Oehmcke, S., Wu, Y., Barnard, K.: Attentional feature fusion. In: 2021 IEEE Winter Conference on Applications of Computer Vision (WACV) (Jan 2021)

\bibitem{Degen2008}
Degen, J., Wegscheid-Gerlach, C., Zaliani, A., Rarey, M.: On the art of compiling and using 'drug-like' chemical fragment spaces. ChemMedChem  \textbf{3}(10),  1503--7 (Oct 2008)

\bibitem{Gao_Fokoue_Luo_Iyengar_Dey_Zhang_2018}
Gao, K.Y., Fokoue, A., Luo, H., Iyengar, A., Dey, S., Zhang, P.: Interpretable drug target prediction using deep neural representation. In: Proceedings of the Twenty-Seventh International Joint Conference on Artificial Intelligence (Jul 2018)

\bibitem{Hua_Song_Feng_Wu_Kittler_Yu_2022}
Hua, Y., Song, X.N., Feng, Z., Wu, X.J., Kittler, J., Yu, D.J.: Cpinformer for efficient and robust compound-protein interaction prediction. IEEE/ACM Transactions on Computational Biology and Bioinformatics p. 1–1 (Jan 2022)

\bibitem{Huang_Xiao_Glass_Sun_2021}
Huang, K., Xiao, C., Glass, L.M., Sun, J.: Moltrans: Molecular interaction transformer for drug target interaction prediction. Bioinformatics p. 830–836 (May 2021)

\bibitem{Jacob2008ProteinligandIP}
Jacob, L., Vert, J.P.: Protein-ligand interaction prediction: an improved chemogenomics approach. Bioinformatics  \textbf{24},  2149 -- 2156 (2008)

\bibitem{10.1371/journal.pcbi.1007129}
Lee, I., Keum, J., Nam, H.: Deepconv-dti: Prediction of drug-target interactions via deep learning with convolution on protein sequences. PLOS Computational Biology  \textbf{15}(6),  1--21 (06 2019)

\bibitem{10.1093/bioinformatics/btac377}
Li, F., Zhang, Z., Guan, J., Zhou, S.: {Effective drug–target interaction prediction with mutual interaction neural network}. Bioinformatics  \textbf{38}(14),  3582--3589 (06 2022)

\bibitem{Liu_Sun_Guan_Zheng_Zhou_2015}
Liu, H., Sun, J., Guan, J., Zheng, J., Zhou, S.: Improving compound–protein interaction prediction by building up highly credible negative samples. Bioinformatics p. i221–i229 (Jun 2015)

\bibitem{10.1093/bioinformatics/btaa921}
Nguyen, T., Le, H., Quinn, T.P., Nguyen, T., Le, T.D., Venkatesh, S.: {GraphDTA: predicting drug–target binding affinity with graph neural networks}. Bioinformatics  \textbf{37}(8),  1140--1147 (10 2020)

\bibitem{pliakos2020drug}
Pliakos, K., Vens, C.: Drug-target interaction prediction with tree-ensemble learning and output space reconstruction. BMC bioinformatics  \textbf{21},  1--11 (2020)

\bibitem{sachdev2019comprehensive}
Sachdev, K., Gupta, M.K.: A comprehensive review of feature based methods for drug target interaction prediction. Journal of biomedical informatics  \textbf{93},  103159 (2019)

\bibitem{sun2020graph}
Sun, M., Zhao, S., Gilvary, C., Elemento, O., Zhou, J., Wang, F.: Graph convolutional networks for computational drug development and discovery. Briefings in bioinformatics  \textbf{21}(3),  919--935 (2020)

\bibitem{Tsubaki_Tomii_Sese_2019}
Tsubaki, M., Tomii, K., Sese, J.: Compound-protein interaction prediction with end-to-end learning of neural networks for graphs and sequences. Bioinformatics p. 309–318 (Jan 2019)

\bibitem{DBLP:journals/corr/abs-1810-00826}
Xu, K., Hu, W., Leskovec, J., Jegelka, S.: How powerful are graph neural networks? CoRR  \textbf{abs/1810.00826} (2018)

\bibitem{zhang2021motif}
Zhang, Z., Liu, Q., Wang, H., Lu, C., Lee, C.K.: Motif-based graph self-supervised learning for molecular property prediction. Advances in Neural Information Processing Systems  \textbf{34},  15870--15882 (2021)

\bibitem{9965612}
Zhao, Q., Duan, G., Zhao, H., Zheng, K., Li, Y., Wang, J.: Gifdti: Prediction of drug-target interactions based on global molecular and intermolecular interaction representation learning. IEEE/ACM Transactions on Computational Biology and Bioinformatics  \textbf{20}(3),  1943--1952 (2023)

\bibitem{Zhao_Yang_Cheng_Li_Wang_2022}
Zhao, Q., Yang, M., Cheng, Z., Li, Y., Wang, J.: Biomedical data and deep learning computational models for predicting compound-protein relations. IEEE/ACM Transactions on Computational Biology and Bioinformatics p. 2092–2110 (Jul 2022)

\bibitem{10.1093/bioinformatics/btab715}
Zhao, Q., Zhao, H., Zheng, K., Wang, J.: {HyperAttentionDTI: improving drug–protein interaction prediction by sequence-based deep learning with attention mechanism}. Bioinformatics  \textbf{38}(3),  655--662 (10 2021)

\bibitem{zitnik2019machine}
Zitnik, M., Nguyen, F., Wang, B., Leskovec, J., Goldenberg, A., Hoffman, M.M.: Machine learning for integrating data in biology and medicine: Principles, practice, and opportunities. Inf Fusion  \textbf{50},  71--91 (Oct 2019), journal Article

\bibitem{10.1093/bioinformatics/bty593}
Öztürk, H., Özgür, A., Ozkirimli, E.: {DeepDTA: deep drug–target binding affinity prediction}. Bioinformatics  \textbf{34}(17),  i821--i829 (09 2018)

\end{thebibliography}
%






\end{document}